\documentclass[conference]{IEEEtran}
\IEEEoverridecommandlockouts

\usepackage{cite}
\usepackage{amsmath,amssymb,amsfonts}
\usepackage{algorithmic}
\usepackage{graphicx}
\usepackage{textcomp}
\usepackage{xcolor}
\usepackage{float}
\def\BibTeX{{\rm B\kern-.05em{\sc i\kern-.025em b}\kern-.08em
    T\kern-.1667em\lower.7ex\hbox{E}\kern-.125emX}}
\usepackage{tabularx}
\usepackage{booktabs}   
\usepackage{multirow}   
\usepackage{hyperref}

\begin{document}

\title{From TF-IDF to Transformers: A Comparative and Ensemble Approach to Sentiment Classification}

\author{
\IEEEauthorblockN{Dip Biswas Shanto}
\IEEEauthorblockA{\textit{School of Computer Engineering} \\
\textit{KIIT Deemed to be University}\\
Bhubaneswar, India \\
shantodipbiswas@gmail.com}
\and
\IEEEauthorblockN{Mitali Yadav}
\IEEEauthorblockA{\textit{School of Computer Engineering} \\
\textit{KIIT Deemed to be University}\\
Bhubaneswar, India \\
miitaliyadav@gmail.com}
\and
\IEEEauthorblockN{Prajwal Panth}
\IEEEauthorblockA{\textit{School of Computer Engineering} \\
\textit{KIIT Deemed to be University}\\
Bhubaneswar, India \\
prajwal.panth21@gmail.com}
\and
\IEEEauthorblockN{Suresh Chandra Satapathy}
\IEEEauthorblockA{\textit{School of Computer Engineering} \\
\textit{KIIT Deemed to be University}\\
Bhubaneswar, India \\
suresh.satapathyfcs@kiit.ac.in}
}

\maketitle
\begin{abstract}
Sentiment analysis, also referred to as opinion mining, primarily tries to extract opinion from any text-based data. In the context of movie reviews and critics, sentimental analysis can be a helpful tool to predict whether a movie review is generally positive or negative. It can be difficult for the ML models to understand the context or metaphysical sentiment accurately, as ML models rely largely on statistical word representations. The objective of this paper is to examine and categorise movie reviews into positive and negative sentiments. Diverse machine learning models are considered in doing so, and Natural Language Processing (NLP) methodologies are employed for data preprocessing and model assessment. The IMDb dataset is used.  Specifically, Naïve Bayes, Logistic Regression, Support Vector Machines (SVM), LightGBM, LSTM, and transformer-based models such as RoBERTa and DistilBERT were evaluated. After a lot of testing with accuracy, precision, recall, F1-score, and ROC-AUC, RoBERTa performed better than all the other models, with an accuracy of 93.02\%. A soft voting ensemble that combined all the models also improved classification performance, showing that model ensembling works well for sentiment analysis.
\end{abstract}


\section{Introduction}
A movie review simply can be defined as a critical analysis of a movie that includes a variety of components such as the storyline, theme, direction, genre, screenplay, acting, etc. The main goal of a movie review is simply to give an insight into the movie's quality. In a text-based movie review, there are a lot of nuances hidden between the lines. ML models and transformer models, such as SVM, Decision Tree, and Neural Network, can be used to decode pure opinions. This paper focuses on the classification of those movie reviews using machine learning models. Some of the models, such as logistic regression, Naive Bayes, Linear SVM, and LightGBM, were trained at the sentence level. Further, they were aggregated to understand review-level sentiment. Comparative analysis determines the most efficient model and aggregation strategy for resilience sentiment classification. At last, to leverage the advantage of diverse models, a soft voting ensemble method was explored that combined three predicted probabilities from all the models used before to improve the overall sentiment classification precision.

The structure of the paper is as follows: Section II presents the related work, Section III outlines the dataset along with the preprocessing methodology. Section IV has addressed the feature extraction and learning models used. Finally, the results are discussed in Section V, and Section VI concludes the paper.

\begin{figure*}[ht!]
    \centering
    \includegraphics[width=0.9\textwidth, height=4.5cm]{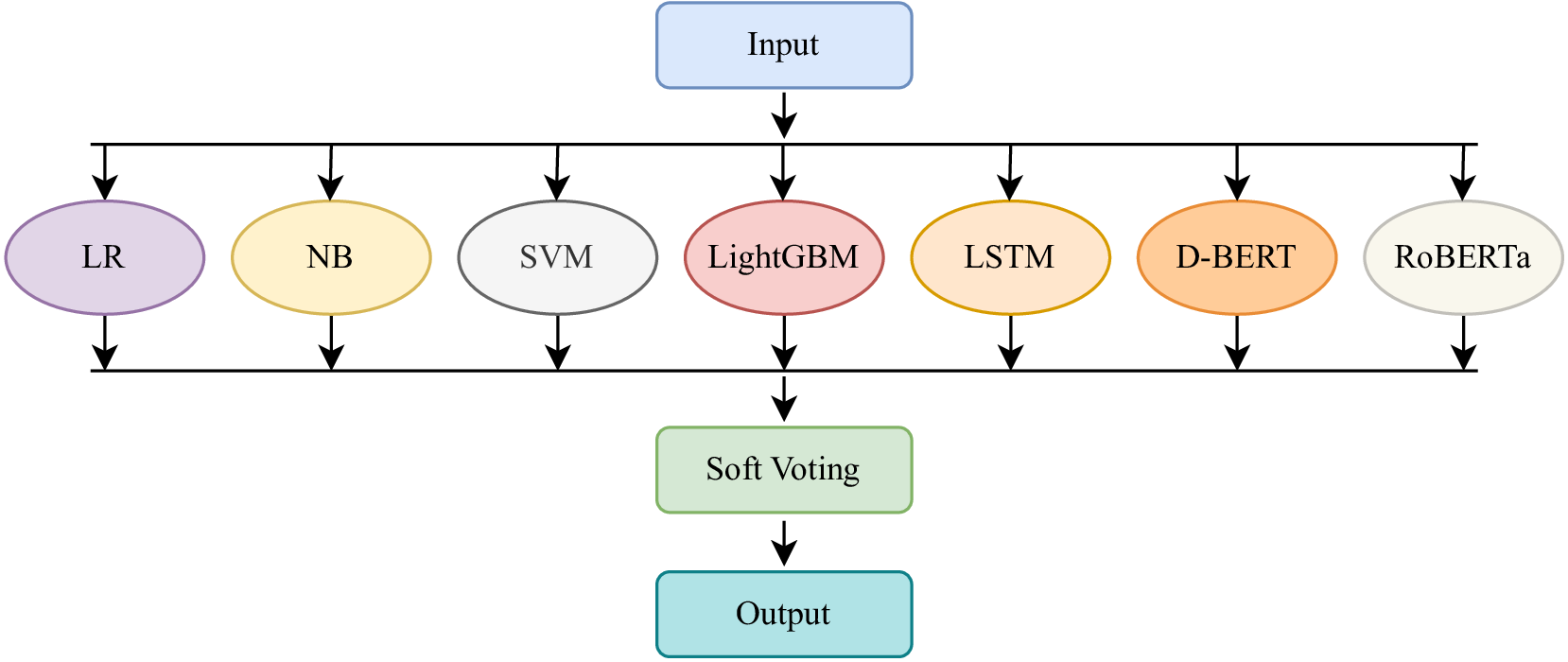}
    \caption{Soft Voting Ensemble Architecture.}
    \label{fig:1}
\end{figure*}

\section{Literature Survey and Related Works}\label{sec2}

Sentiment analysis has a long history, from counting the number of words and using dictionaries to look up to now using deep learning models or transformer-based models.  AlShammari proved that TF-IDF still pulls its weight for quick keyword extraction in Python\cite{alshammari2023}. Kibriya and co. pushed Multinomial  Naive Bayes as a solid, simple, yet fast and reliable way of sentiment analysis when a person has a lot of text\cite{kibriya2005}. Then came hybrid approaches that used both lexical information and machine learning models. This improved the sentimental classification. The IMDb dataset was used by Kumar et al. He made a framework that combined both machine learning techniques and lexical-based features for sentiment analysis. Kumar's work showed that it is better to use both than to rely purely on either lexical or statistical approaches\cite{kumar2019}. Deep learning techniques have enhanced sentiment analysis by identifying patterns within the text. Yu investigated the application of CNN, Convolutional Neural Network, with TF-IDF and Bag of Words representation for the classification of movie reviews. This demonstrated that a convolutional model can be used to detect local semantic patterns when supported with a robust processing method.
Meanwhile, researchers have also been looking into cluster-based methods. to analyse the sentiments in IMDB reviews. This can be used as a testimony that unsupervised techniques could still be useful for text analysis.\cite{topal2016}

\section{Dataset and Preprocessing}\label{sec3}
\subsection{Dataset}

For the dataset, the widely recognized IMDb dataset \cite{imdb} was used, that compromised 50,000 movie reviews. The reviews were then split in half between training and test sets. Each set is balanced, 25,000 labelled positive and 25,000 labelled negative. The dataset had no missing or mismatched entries. The sheer size of the dataset made the testbed reliable for machine learning, at the same time making sure that there was a good mix of positive and negative sentiment classes. This made the classification model plausible to be tested in a strong way.

\begin{table}[h]
\centering
\caption{Summary Statistics of the IMDB Movie Reviews Dataset}
\label{tab:dataset}
\renewcommand{\arraystretch}{1.25} 
\setlength{\tabcolsep}{8pt} 
\begin{tabular}{lcc}
\toprule
\textbf{Column} & \textbf{Metric} & \textbf{Value} \\
\midrule
\multirow{5}{*}{\textbf{Review}} 
    & Valid Entries     & 50,000 (100\%) \\
    & Mismatched        & 0 \\
    & Missing           & 0 \\
    & Unique Values     & 49,582 \\
    & Most Common Value & ``Loved today..'' \\
\midrule
\multirow{5}{*}{\textbf{Sentiment}} 
    & Valid Entries     & 50,000 (100\%) \\
    & Mismatched        & 0 \\
    & Missing           & 0 \\
    & Unique Values     & 2 (Positive, Negative) \\
    & Most Common Value & Positive (50\%) \\
\bottomrule
\end{tabular}
\end{table}

\subsection{Data preprocessing}
Data preprocessing converts raw text into a structured format, ensuring that the dataset is free from inconsistency and irrelevant information, which could negatively impact the overall performance of our model. The first step in data preprocessing for this study was text normalisation. It was followed by tokenisation, which breaks sentences down into their individual words. Then, stemming or lemmatisation processes. 
This process simplifies words to their most basic forms. To make sure the test is unbiased and fair, the dataset is divided into two parts: a training set and a testing set. The TF-IDF vectorizer also converts text into numbers by capturing the significance of words based on their frequency and relevance. These preprocessing steps work together to improve the model's ability to detect relevant text patterns, which in turn improves overall sentiment classification.
\section{Proposed Methodology}\label{sec4}
\begin{figure*}[ht!]
    \centering    \includegraphics[width=\textwidth, height=6cm]{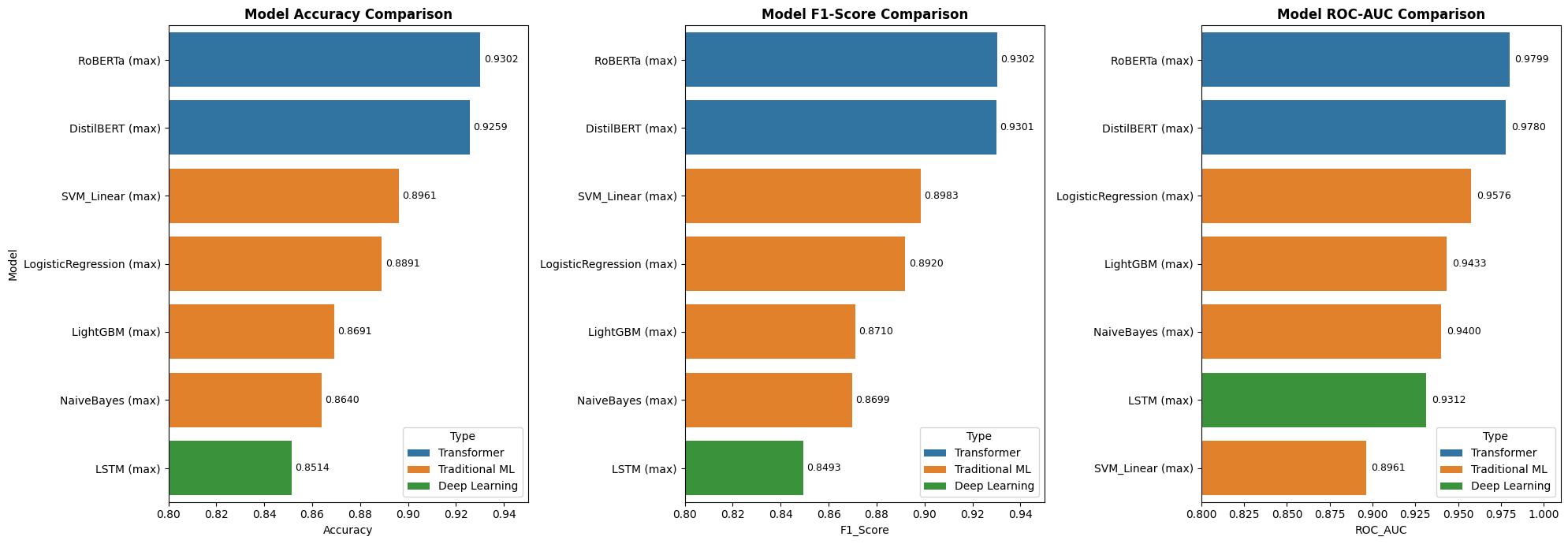} 
    \caption{Evaluation of Model Performance Metrics: Accuracy, F1-Score, and ROC-AUC.}
    \label{fig:2}
\end{figure*}

\subsection{Feature Engineering}
Two main strategies were used to turn words into numbers. For the traditional models, TF-IDF, capping the vocabulary to the 10,000 most informative words, making sure not to leak any information from the test set was used.\cite{das2018tfidf}. The vectorizer was fitted on the training corpus and applied consistently to the training and test data, ensuring that inverse document frequencies were calculated only from the training data to prevent leakage.  

Meanwhile, for transformer models, subword tokenisation was considered and used their contextual embeddings \cite{bhargavi2025comparative}. providing richer semantic representation, more dynamic representations straight from models like RoBERTa and DistilBERT compared to static TF–IDF features.

\subsection{Model Training and Architecture}

\textbf{SVM (Linear Kernel):}  
Support Vector Machine (SVM) was used to train on TF-IDF features. As SVM handles high dimensionality and sparsity of feature space, a linear kernel was considered. 
Performance: Accuracy = 0.8961, F1 = 0.8983.  

\textbf{DistilBERT:}  
DistilBERT is a compact variant of BERT. It was fine-tuned for sentiment classification. To do so, a softmax classification was added over the pretrained encoder.\cite{kaur2025transformers}. Fine-tuning adapted contextual embeddings for the sentiment dataset while maintaining computational efficiency.  
Performance: Accuracy = 0.9259, F1 = 0.9301. 

\textbf{RoBERTa:}  
Meanwhile, RoBERTa is an optimized variant of BERT. It was also fine-tuned for sentiment analysis to improve the overall contextual learning of the data by removal of next sentence predictor, large batch sizes, dynamic masking, and exposure to significantly more data, enabling superior complex contextual learning. \cite{liu2019roberta}.   
Performance: Accuracy = 0.9302, F1 = 0.9302.

\subsection{Soft Voting Ensemble}

To increase the overall accuracy, an ensemble approach was considered. Soft voting, an ensemble approach, where each model votes by averaging its probability prediction, became beneficial and increased the accuracy significantly. The lineup: logistic regression, Naive Bayes, SVM, LightGBM, LSTM, DistilBERT, and RoBERTa. Each model was given equal weight. Then the ensemble picked the class with the highest combined score.

This methodology really shone by resolving disagreements between models and by handling tricky reviews. By averaging things out, the ensemble reduces overall variance and tends to improve the result across all metrics compared to using RoBERTa alone.

For each input review $\mathbf{x}$, each model $m_i$ produces a probability distribution over the classes as defined in Equation (1):
\begin{equation}
P_m(c|\mathbf{x}) = [p_{m,1}, p_{m,2}, \dots, p_{m,C}]
\end{equation}

The ensemble combines these probabilities through weighted averaging as shown in Equation (2):

\noindent
\begin{equation}
P_{\text{ensemble}}(c|\mathbf{x}) = \frac{1}{N}\sum_{i=1}^{N} w_i \cdot P_{m_i}(c|\mathbf{x})
\end{equation}

where $w_i$ represents the weight assigned to model $m_i$, with $\sum_{i=1}^{N} w_i = 1$. In this implementation, equal weighting was used($w_i = \frac{1}{N}$) for all models.

The final sentiment prediction is determined in Equation (3):

\noindent
\begin{equation}
\hat{y} = \arg\max_{c \in C} P_{\text{ensemble}}(c|\mathbf{x})
\end{equation}

All models were trained with carefully selected hyperparameters to ensure stable convergence and fair comparison. For traditional machine learning models, TF–IDF features were generated with a maximum vocabulary size of 10,000 and an n-gram range of 1–3. Logistic Regression used the saga optimiser with 1000 iterations, Multinomial Naïve Bayes used additive smoothing, and the linear SVM was trained with 2000 iterations. LightGBM was configured with 150 estimators, a learning rate of 0.05, and 63 leaves per tree.

For the deep learning approach, a bidirectional LSTM model with 64-dimensional embeddings and 64 LSTM units was trained using the Adam optimiser with binary cross-entropy loss for four epochs and a batch size of 128.

Transformer models, including DistilBERT and RoBERTa, were fine-tuned using the Hugging Face Transformers framework with a maximum sequence length of 128 tokens, a batch size of 32, weight decay of 0.01, and 500 warmup steps.
\textit{\textbf{Implementation Details:}}
The ensemble model incorporated seven diverse classifiers: SVM, Logistic Regression, Naïve Bayes, LightGBM, LSTM, DistilBERT, and RoBERTa. Having diverse models made the ensemble especially strong and successful:  

\begin{itemize}
\item \textbf{Traditional models (SVM, LR, NB, LightGBM)}: Captured lexical patterns and word-level features through TF-IDF representations
\item \textbf{Neural models (LSTM)}: Captured sequential dependencies and contextual information
\item \textbf{Transformer models (DistilBERT, RoBERTa)}: Leveraged contextual embeddings and self-attention mechanisms for nuanced understanding
\end{itemize}

The approach of soft voting demonstrated effectiveness by handling the individual models when they showed disagreement, i.e. it handled the ambiguous cases particularly well. The mechanism was averaging the probability estimates, which helped in reducing the variance and mitigating the individual biases of a model. Hence, it resulted in 1-3\%improvement across all metrics compared to using a single best model, which was RoBERTa.

\subsection{Model Explainability with SHAP}
To achieve Model transparency SHapley Additive exPlanations (SHAP) was used\cite{lundberg2017shap, kokalj2021bertshap}. The mechanism that SHAP follows is simply to assign each feature a SHAP value, which indicates its marginal contribution to the prediction. In terms of SVM, interpretability was straightforward. That is, SAP values corresponded to the product of TF-IDF feature weights and SVM coefficients. Words with strong positive coefficients, such as good, better, excellent, etc., contributed to positive prediction. At the same time, negative coefficients, for example, boring, bad, detrimental, automatically contributed to negative prediction.

Then, for the transformer model, SHAP was applied using perturbation-based techniques \cite{khan2025shap}, which means the input sentence was systematically masked. Then, the fine-tuned models were re-evaluated, and token-level attributions were calculated. For visualisation, such as force plots and bar charts, the influential tokens, that is, positive contributions, were colored in red and negative contributions were colored in blue. 

\section{Results Analysis}\label{sec5}
Model performance was evaluated on the test dataset based on \textbf{accuracy} and \textbf{F1-score}. The corresponding results reported in Table~\ref{tab:results}.

\begin{table}[h!]
\centering
\caption{Performance of Sentiment Analysis Models}
\label{tab:results}
\renewcommand{\arraystretch}{1.25} 
\setlength{\tabcolsep}{10pt}       
\begin{tabular}{lccc}
\toprule
\textbf{Model} & \textbf{Accuracy} & \textbf{F1-Score} & \textbf{ROC AUC} \\
\midrule
RoBERTa            & 0.9302 & 0.9302 & 0.9799 \\
DistilBERT         & 0.9259 & 0.9301 & 0.9780 \\
SVM                & 0.8961 & 0.8983 & 0.8961 \\
Logistic Regression & 0.8891 & 0.8920 & 0.9576 \\
LightGBM           & 0.8691 & 0.8701 & 0.9433 \\
Naïve Bayes        & 0.8640 & 0.8699 & 0.9400 \\
LSTM               & 0.8515 & 0.8493 & 0.9312 \\
\bottomrule
\end{tabular}
\end{table}

\subsection{Ablation Study}

A semantic evaluation study was conducted in order to quantify the contribution of every individual component in the pipeline of semantic analysis. It examined pre-processing techniques, feature extraction models, and model-specific components.
\begin{figure}[ht!]
    \centering
    \begin{minipage}[]{0.49\textwidth} 
        \centering
        \includegraphics[width=\textwidth, height=3.5cm]{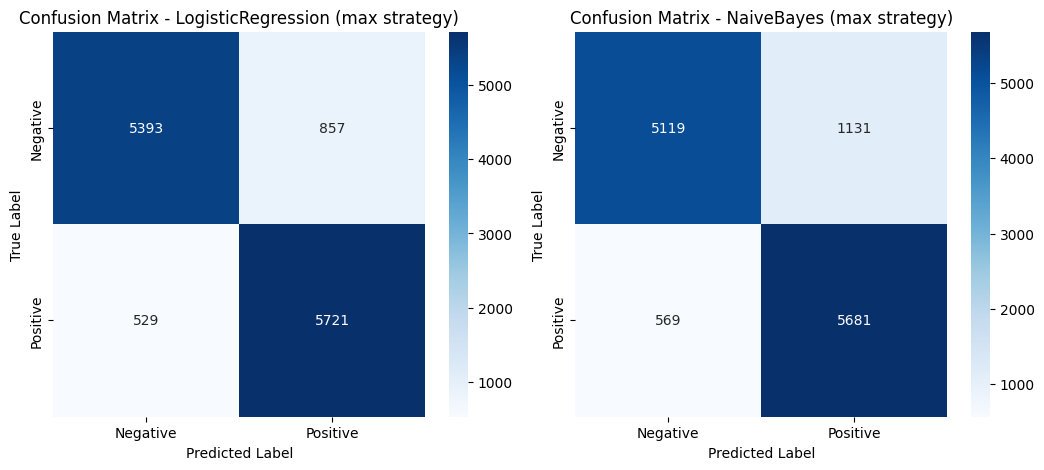}
        \caption{Confusion Matrices for Logistic Regression and Naïve Bayes.}
        \label{fig:3}
    \end{minipage}\hfill
\end{figure}

\begin{figure}[ht!]
    \centering
    \begin{minipage}[]{0.49\textwidth} 
        \centering
        \includegraphics[width=\textwidth, height=3.5cm]{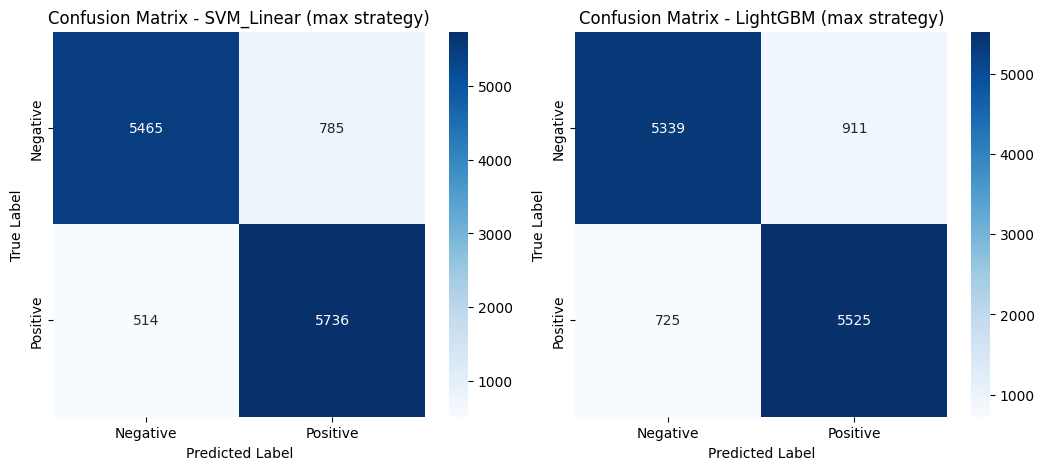}
        \caption{Confusion Matrices for LightGBM and SVM models.}
        \label{fig:4}
    \end{minipage}\hfill
\end{figure}

\begin{figure}[ht!]
    \centering
    \begin{minipage}[]{0.3\textwidth} 
        \centering
        \includegraphics[width=\textwidth, height=3.5cm]{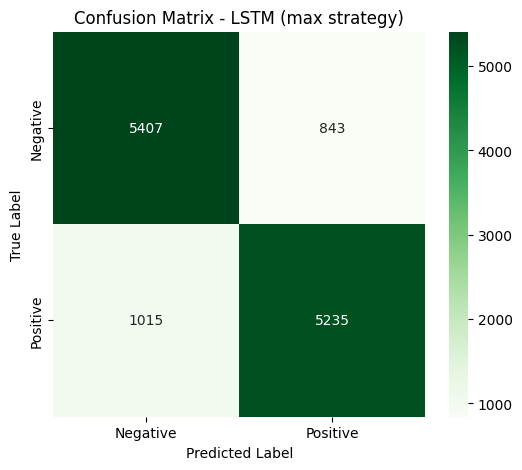}
        \caption{Confusion matrix of LSTM model.}
        \label{fig:5}
    \end{minipage}\hfill
\end{figure}

\begin{figure}[ht!]
    \centering
    \begin{minipage}[]{0.49\textwidth} 
        \centering
        \includegraphics[width=\textwidth, height=3.5cm]{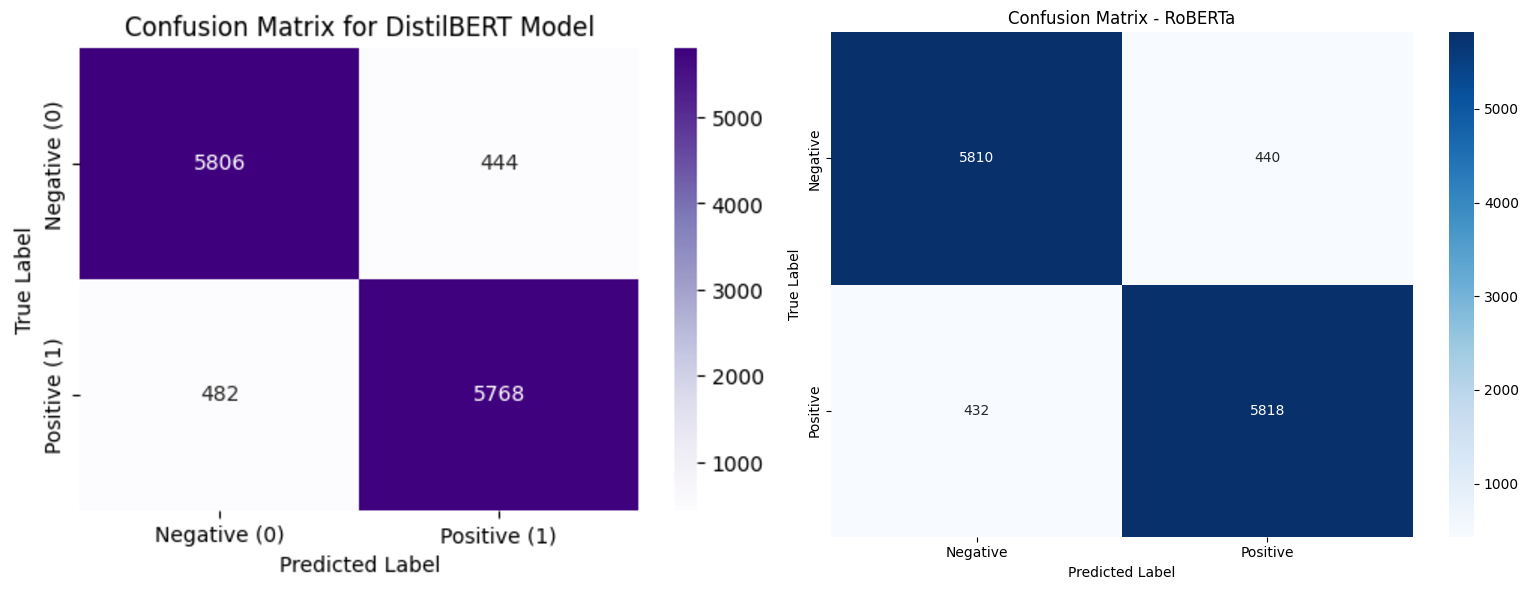}
        \caption{Confusion Matrices for DistilBERT and RoBERTa models.}
        \label{fig:6}
    \end{minipage}\hfill
\end{figure}

\begin{figure*}
        \centering
        \includegraphics[width=0.8\textwidth, height=8cm]{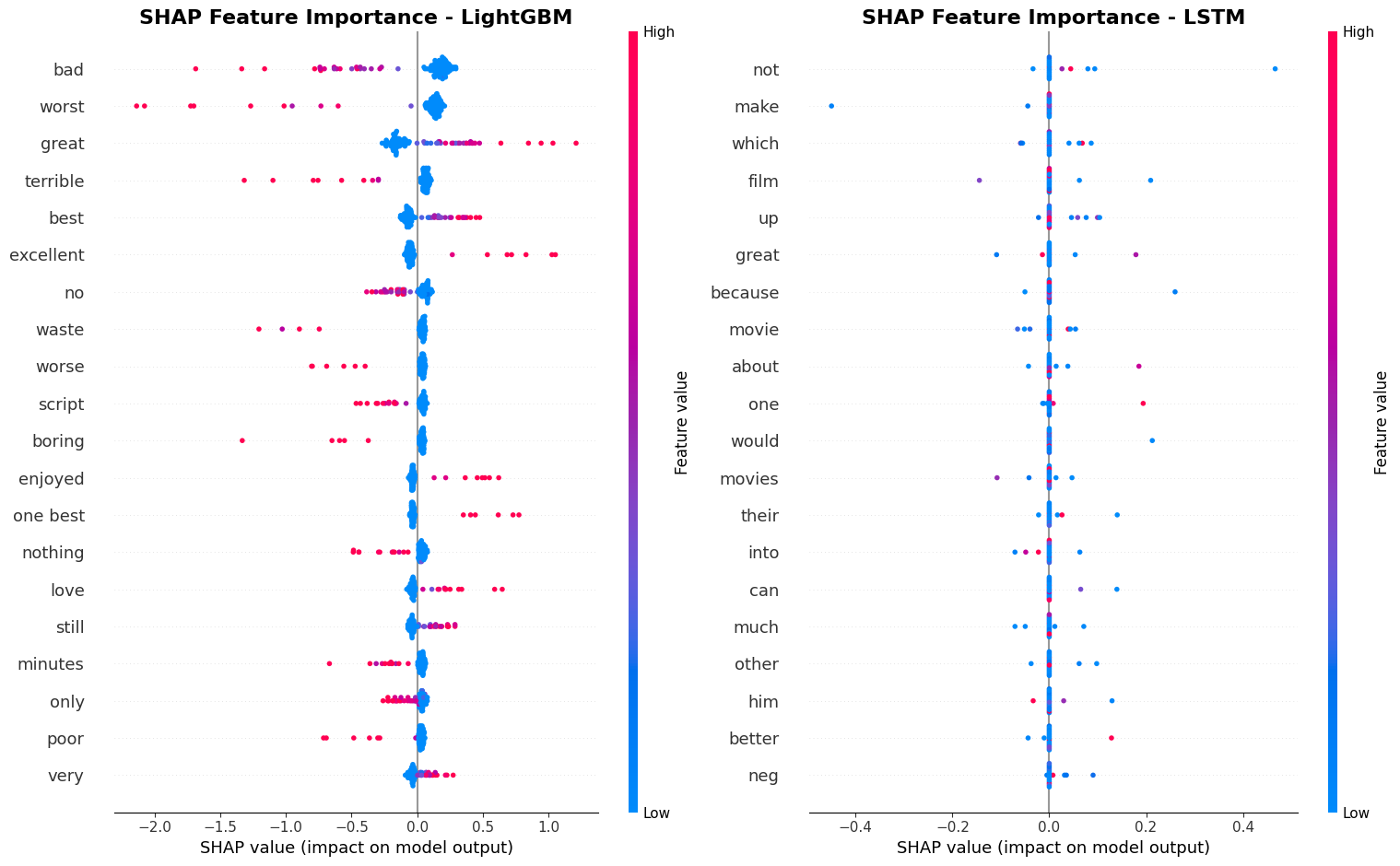}
        \caption{SHAP feature importance for LightGBM and LSTM models.}
        \label{fig:7}
\end{figure*}

\begin{figure}[ht!]
    \centering
    \begin{minipage}[]{0.3\textwidth} 
        \centering
        \includegraphics[width=\textwidth, height=4 cm]{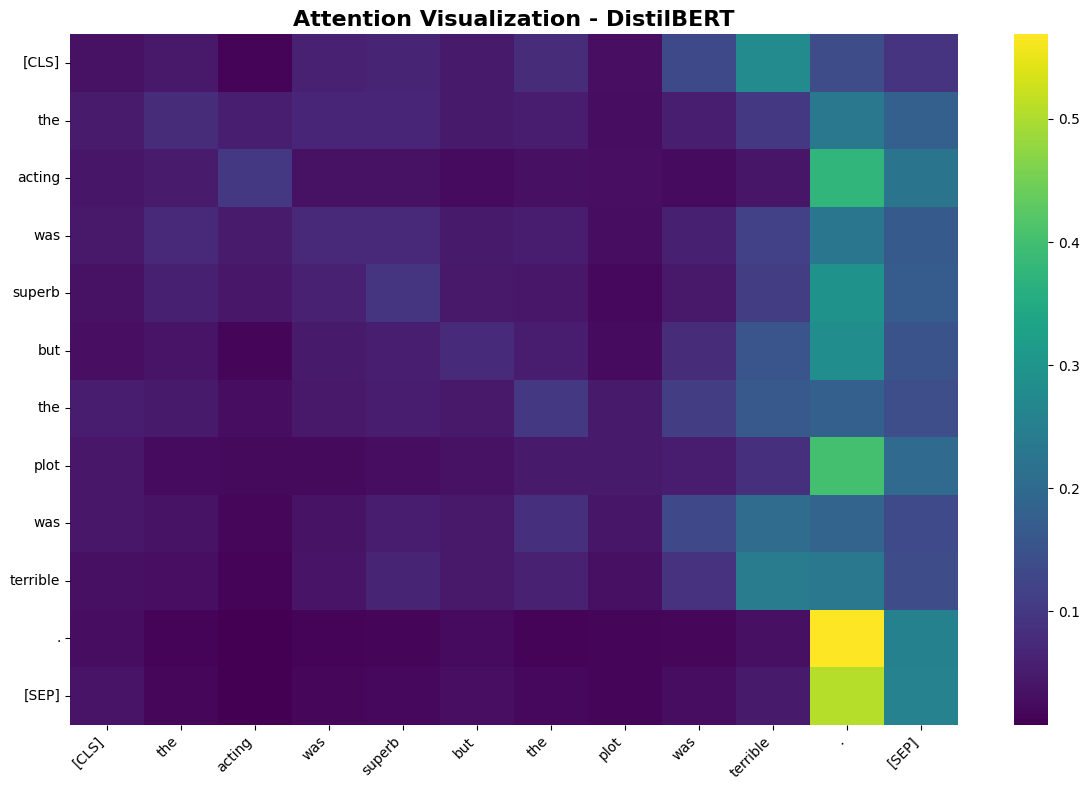}
        \caption{Attention visualization for DistilBERT model.}
        \label{fig:8}
    \end{minipage}\hfill
\end{figure}

\begin{figure}[ht!]
    \centering
    \begin{minipage}[]{0.3\textwidth} 
        \centering
        \includegraphics[width=\textwidth, height=4 cm]{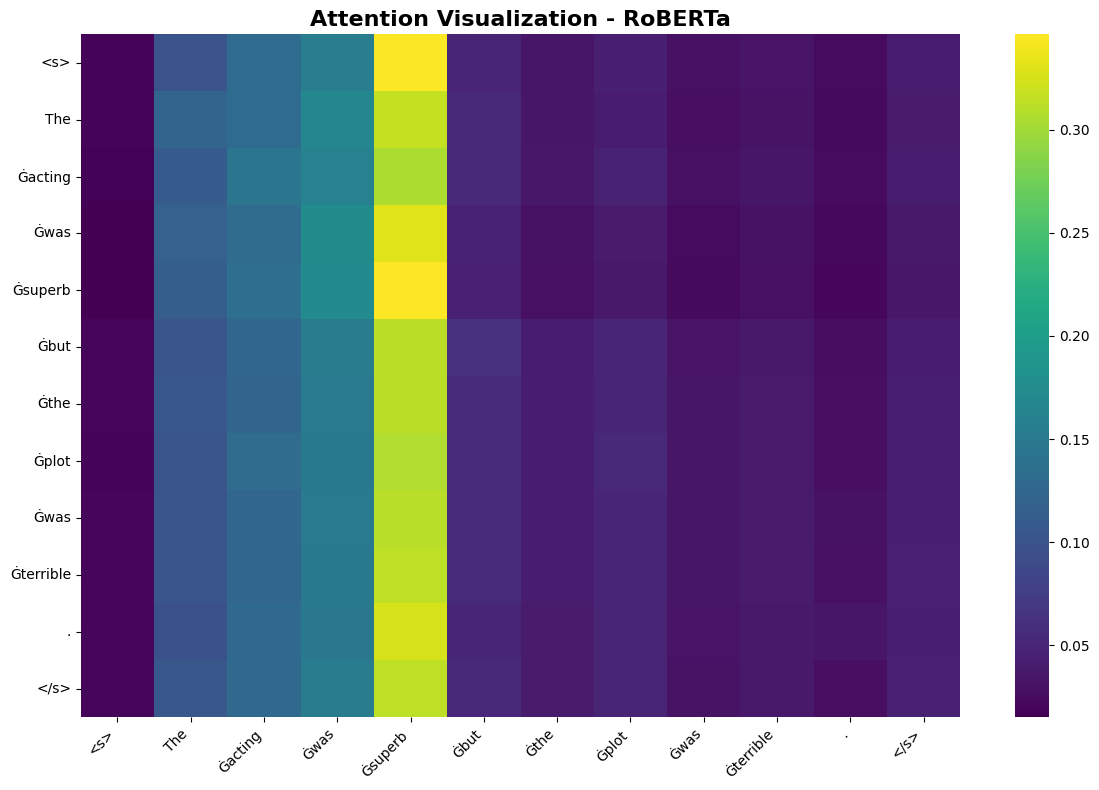}
        \caption{Attention visualization for RoBERTa model.}
        \label{fig:9}
    \end{minipage}\hfill
\end{figure}

\textbf{Preprocessing Impact:} 
The advanced negation handling mechanism yielded the most substantial performance gain among traditional machine learning models, improving F1-scores by approximately \textbf{+3.2--4.1\%}. 

\medskip

\textbf{Feature Extraction Analysis:} 
Incorporating \textbf{TF-IDF with n-gram ranges (1--3)} outperformed the unigram-based approach, resulting in an improvement of about \textbf{3.7--4.0\%} in overall accuracy. 

\medskip

\textbf{Model-Specific Components:} 
Within transformer-based architectures, \textbf{dynamic masking during RoBERTa fine-tuning} contributed an additional \textbf{+2.1\%} increase in accuracy, highlighting the benefit of adaptive context exposure. 

\medskip

\textbf{Ensemble Contribution:} 
The \textbf{soft voting ensemble} consistently enhanced performance across all error categories, particularly for complex or sarcastic reviews, where individual models exhibited complementary strengths. Notably, the ensemble reduced misclassification rates for sarcastic reviews by approximately \textbf{23\%}.

\begin{table}[!ht]
\centering
\caption{Key Ablation Study Results (Performance Impact)}
\label{tab:ablation}
\renewcommand{\arraystretch}{1.15} 
\setlength{\tabcolsep}{8pt}        
\scriptsize                        
\begin{tabular}{lcc}
\toprule
\textbf{Component} & \textbf{Traditional Models} & \textbf{Transformer Models} \\
\midrule
Negation Handling         & +3.2--4.1\% F1  & +1--2\% F1 \\
N-gram Features (1--3)    & +3.7--4.0\% Acc & N/A \\
RoBERTa                   & N/A             & +2.1\% Acc \\
Learning Rate Scheduling  & N/A             & +1.8\% Acc \\
Ensemble Voting           & +1--3\% All metrics & +1--2\% \\
\bottomrule
\end{tabular}
\end{table}

\subsection{Error Analysis}

Although the models performed well with good accuracy, some reviews were still misclassified. Analyzing the errors revealed a systematic pattern across all models. Some of the challenges included: :

\textbf{Complex Negations:}Double negation and cross-casual negation were difficult to be classified, hence were often misclassified due to its complicated nature. For example, the phrase ``not without its advantage" was often categorized as negative despite expressing positive sentiment.

\textbf{Sarcasm and Irony:} Models demonstrated limited capabilities when it came to detecting sarcasm, as they relied largely on the literal interpretation of the review.Reviews containing phrases like ``Just what I needed, another terribly written superhero movie'' were consistently misclassified.

\textbf{Mixed Sentiments:}  Some review content includes both good and bad things. When the final sentiment is not clear in the review itself, it can be a challenge for the model to interpret whether the review is considered positive sentiment or negative sentiment. In such a context, models tend to give more weight to negative expressions.

\textbf{Contextual Limitations:} Traditional TF-IDF models struggle with vocabulary limitations and vast context-dependent meaning. Transformer models often overfitted to specific syntactic patterns, which showed a reduced performance on short reviews ($<$50 words).

\textbf{Quantitative Patterns:} Error rates were significantly high when the review comprised sarcasm (18.9\% error rate) and complex negations (16.8\% error rate) was present, as already discussed above. A standard review was easier to classify as positive or negative sentiment (7.9\% error rate). However, the soft voting ensemble reduced the error rates across all the categories as it leveraged the complementary strengths of different model architectures.

\subsection{Observations and Insights}
Discussing about observation and insights in this study, transformer-based models, i.e, RoBERTa achieved the highest accuracy along with the highest F1 score, which confirmed that the ability to capture contextual dependency and subtle semantic nuances in text is higher in transformer-based model~\cite{pookduang2025roberta}. SVM that used TF-IDF features also performed competitively. This indicated that the textual sentiment signals are often linearly separable in high-dimensional space~\cite{huzyan2023SVM}. The DistilBERT maintained a strong performance with limited computational power. Hence, can be considered a better alternative while being computationally efficient.

In SVM, feature weight and TF-IDF values highlighted key positive and negative semantic words\cite{ribeiro2016trust}. For transformer model, permutation-based SHAP provided token-level attribution. It confirmed that meaningful words such as such as \emph{excellent}, \emph{captivating}, and even \emph{boring} contributed effectively to the prediction

The comparative analysis indicated that the transformer models were better at semantic patterns, and SVM offered an efficient and interpretable benchmark. Conventional models, including Naïve Bayes and lightGBM, depend on the TF-IDF features. It may encounter difficulties with contextual dependency that includes sarcasm or negation. Due to training on limited data, the LSTM model exhibited inferior performance. Transformer-based models, as expected, like DistilBERT and RoBERTa, attained superior accuracy. It did so by using Transfer Learning and self-attention that helped capture contextual relations better. Hence, it can be established that contextual language models are efficient for complicated sentiment analysis tasks. At the same time, traditional methodology can remain of value when computational efficiency is limited.

\section{Conclusion}\label{sec6}
This study has offered a comparative analysis of traditional machine learning, deep learning, and transformer-based models. The sentimental classification used the IMDB movie review dataset. Among all the trained models, ROBERT was considered the highest-performing model. Soft voting ensemble can be viewed as a beneficial approach to enhance the overall classification accuracy, which highlights the advantages of using a model ensemble. The experimental result produced by this study showed that the transformer-based model performed better, as it captured contextual relations and also understood semantic nuances better than traditional approaches did. DistilBERT can be considered as an effective alternative that has strong representational power while being computationally efficient. SVM, along with TF-IDF features, is still considered a very strong baseline. This is because it is easy to understand, and it also does not require a lot of computing power. The use of SHAP helped to explain the prediction at the token level, making the model comparatively more transparent. To sum up, the results acquired showed that transformer-based models are better suited for complex and complicated tasks. Sentiment analysis task. The fact that the traditional model learning method was still helpful with limited computing power should not be overlooked. For future research, hybrid models can be explored further at the same time, and multilingual data sets and domain-specific fine-tuning might be considered beneficial to improve sentiment classification even further.

\bibliographystyle{IEEEtran}
\bibliography{references}
\end{document}